\documentclass[letterpaper, 10 pt, conference]{ieeeconf}  

\IEEEoverridecommandlockouts                              

\usepackage[pdftex]{graphicx}
\usepackage{cite}
\usepackage{amsmath, amssymb}
\usepackage{array}
\usepackage{url}
\usepackage{xcolor}
\usepackage{blindtext}
\usepackage{booktabs}
\usepackage[section]{placeins}
\usepackage{subcaption}
\usepackage{multirow}
\usepackage{float}

\title{\LARGE \bf
Adversarial Mixture Density Networks: Learning to Drive Safely from Collision Data
}

\author{Sampo Kuutti,
	Saber Fallah, 
	\thanks{Sampo Kuutti and Saber Fallah are with the Connected and Autonomous Vehicles Lab, University of Surrey, Guildford, GU2 7XH, UK. Email: \{s.j.kuutti, s.fallah\}@surrey.ac.uk}
	Richard Bowden
	\thanks{Richard Bowden is with the Centre for Vision, Speech and Signal Processing, University of Surrey, GU2 7XH, UK. Email: r.bowden@surrey.ac.uk}
} 

\begin{document}

\maketitle
\thispagestyle{empty}
\pagestyle{empty}

\begin{abstract}
Imitation learning has been widely used to learn control policies for autonomous driving based on pre-recorded data. However, imitation learning based policies have been shown to be susceptible to compounding errors when encountering states outside of the training distribution. Further, these agents have been demonstrated to be easily exploitable by adversarial road users aiming to create collisions. To overcome these shortcomings, we introduce Adversarial Mixture Density Networks (AMDN), which learns two distributions from separate datasets. The first is a distribution of safe actions learned from a dataset of naturalistic human driving. The second is a distribution representing unsafe actions likely to lead to collision, learned from a dataset of collisions. During training, we leverage these two distributions to provide an additional loss based on the similarity of the two distributions. By penalising the safe action distribution based on its similarity to the unsafe action distribution when training on the collision dataset, a more robust and safe control policy is obtained. We demonstrate the proposed AMDN approach in a vehicle following use-case, and evaluate under naturalistic and adversarial testing environments. We show that despite its simplicity, AMDN provides significant benefits for the safety of the learned control policy, when compared to pure imitation learning or standard mixture density network approaches.
\end{abstract}

\section{Introduction}
Autonomous vehicles have received increasing interest as a potential solution to transportation issues such as traffic congestion, pollution, and vehicle safety \cite{eskandarian2019research, eskandarian2012handbook, thrun2010toward}. One of the enabling technologies gaining popularity for autonomous driving is deep learning \cite{kuutti2020survey}. Deep learning has been demonstrated to learn driving rules from recorded data, which provide general driving policies that generalise to a wide variety of driving scenarios and outperform traditional approaches. 

A common strategy for learning autonomous driving policies is imitation learning, where a deep neural network is trained to imitate an expert driver using a dataset of demonstrated driving \cite{hussein2017imitation}. Imitation learning has shown to scale well to large amounts of driving data, and to perform well when deployed in similar scenarios to its training data \cite{pomerleau1989alvinn, bansal2018chauffeurnet, wang2018deep, hecker2018end}. However, one of the main limitations of imitation learning is that during training, the states observed by the agent are not dependent on the learned control policy. However, once deployed, the states visited by the agent depend on its own control policy, often leading to compounding errors as the agent visits states outside of its training distribution \cite{bagnell2015invitation, osa2018algorithmic, codevilla2018offline, ross2011reduction}. As the agent experiences events further away from the safe states seen in its training data, the errors in its prediction grow larger, creating a cascading failure. Further issues in imitation learning models can be caused by unpredictable road users \cite{kuutti2020training} or causal confusion \cite{de2019causal}.

A number of solutions have been offered to mitigate these drawbacks in imitation learning. For instance, Dataset Aggregation (DAgger) \cite{ross2011reduction}, first learns from a dataset of expert demonstration after which the agent is deployed in the test environment. During testing, the expert labels each state seen by the agent with the correct action, and in this way, labels for states visited by the agent not seen in the original training data are collected. However, one of the main limitations of this approach is that it requires the expert to be available for labelling large amounts of new data, which is not always possible. Moreover, for a human annotator it may be difficult to label the data when they were not in control of the vehicle (e.g. labelling the correct control action may be difficult for a human when presented with only a still image frame, without further context). An alternative approach that does not query the expert during training is Generative Adversarial Imitation Learning (GAIL) \cite{ho2016generative}, inspired by Generative Adversarial Networks \cite{goodfellow2014generative}. GAIL uses a discriminator during training, which aims to classify trajectories as either coming from the expert or the learned model. Therefore, as training progresses the trajectories by the learned model become more human-like, making the classification more difficult for the discriminator. Other approaches have turned to Reinforcement Learning (RL) instead, which learns from interaction with its environment \cite{sallab2017deep, puccetti2019actor, chae2017autonomous, kuutti2019end}. As the states visited by the RL agent during training are dependent on its own policy, RL tends to provide a more robust policy which can generalise well to different scenarios. However, RL requires a significant amount of training, due to the low sample efficiency of RL algorithms \cite{henderson2017deep}. This typically leads to the need for high fidelity simulators, and restricts the use of RL within real-world use-cases.

In this work, we present Adversarial Mixture Density Networks (AMDN) which provide a method for imitation learning with improved robustness to distribution shift and increased safety in dangerous driving scenarios. The proposed techniques utilises two datasets, one with safe expert driving trajectories with labelled actions and another with trajectories that lead to collisions without labels for correct actions. The AMDN model aims to simultaneously learn two distributions representing both safe and unsafe actions. By learning these two distributions, the agent can then penalise the safe action distribution whenever it becomes too similar to the unsafe action distribution. Using these two distributions, and a measure of similarity between them, the model can then further train the safe action distribution based on the safety-critical scenarios in the collision dataset, without any need for labelling of this dataset. This is similar to Linear Discriminant Analysis (LDA) \cite{fisher1938statistical, fukunaga2013introduction}, which aims to separate classes of data by modelling them as Gaussian distributions, and maximising the mean distance between each class \cite{martinez2001pca}. However, our approach differs significantly from LDA, since in our case the aim is to use regression to infer continuous control actions rather than classify data or reduce the dimensionality of the data. And since the aim is to learn a safer control policy, we only penalise the safe action distribution based on its similarity to the unsafe distribution. Furthermore, by leveraging the non-linearity of deep neural networks, the proposed technique can be trained on more complex data and use-cases. Therefore, our proposed technique extends standard imitation learning and mixture density network approaches, by providing an additional loss during imitation learning, which increases the safety of the learned driving policy and boosts the model's robustness to adversarial agents. We demonstrate the proposed approach in a vehicle following use-case, and evaluate the control policies both in naturalistic driving with typical leading vehicles, as well as in safety-critical environments with adversarial agents who are attempting to create rear-end collisions. We demonstrate that the proposed Adversarial Mixture Density Networks improves the model's robustness in these safety-critical scenarios compared to normal imitation learning and mixture density networks.

The remainder of the paper is as follows. Section II presents the required background and introduces the proposed Adversarial Mixture Density Network technique. Section III demonstrates the effectiveness of the approach in simulated driving tests. Finally, conclusions are presented in Section IV.

\section{Methodology}
We demonstrate the proposed approach in a highway vehicle following use-case. The aim of the model is to control the longitudinal actions of an autonomous vehicle, whilst following the lead vehicle at a safe distance. The model does this by observing states from the vehicle's Radar and inertial sensors, and then infers the correct actions to control the gas and brake pedals of the host vehicle. We give a brief overview of the background on Imitation Learning and Mixture Density Networks in sub-sections II-A and II-B, respectively. We then introduce the proposed Adversarial Mixture Density Networks in sub-section II-C.

\subsection{Imitation Learning}
Imitation Learning (IL) is a type of supervised learning strategy, where an agent learns from expert demonstration of the desired task. The aim is to learn to imitate the expert's behaviour \cite{pomerleau1991efficient}. To do this, an expert demonstration dataset $\mathcal{D}^e$, consisting of observed states $s_t$ and the corresponding actions taken by the expert $\hat{a}_t$, is used. The IL policy denoted by $\pi^{IL}$ and represented by parameters $\theta^{IL}$, aims to find optimal parameters $\theta^*$, by minimising an imitation loss $\mathcal{L}^{IL}$ based on the distance of its predicted action $a_t$ to the expert's demonstrated action $\hat{a}_t$ for the same observed state $s_t$:
\begin{equation}
\theta^* = \arg \min_{\theta^{IL}} \sum_{t}\mathcal{L}^{IL}(\pi^{IL}(s^{IL}_t|\theta^{IL}), \hat{a}_t)
\end{equation}

\subsection{Mixture Density Networks}
A Mixture Density Network (MDN) \cite{bishop1994mixture} is a type of neural network, which can model a distribution as a mixture of parametric distributions: 
\begin{equation}
p(y|x) = \sum_{i = 1}^{M}\alpha_i(x)\phi(y|\theta_i)
\end{equation}
where $y$ is the output, $x$ is the input, $M$ is the number of mixture components, $\alpha_i$ is the mixing coefficient, and $\phi$ is a parametric distribution with parameters $\theta_i$. For instance, a Gaussian distribution $\mathcal{N}(y|x)$ can be represented by its mean and (co-)variance $\theta_i = (\mu_i, \sigma^2_i)$.

Mixture Density Networks can be used to estimate any given density function to an arbitrary accuracy given sufficient number of mixtures \cite{mclachlan1988mixture}. MDNs have been used to model data distributions thanks to their ability to represent multi-modal data distributions and provide uncertainty estimations in various tasks such as speech synthesis \cite{wang2017autoregressive}, future prediction \cite{makansi2019overcoming}, and autonomous driving \cite{hu2018probabilistic, choi2018uncertainty}.

\subsection{Adversarial Mixture Density Networks}
Our proposed Adversarial Mixture Density Network is similar to MDNs in that it also aims to represent parametric distributions given an input $x$. However, instead of learning a mixture of distributions that maps the probability of output $y$ for a given $x$, the AMDN aims to learn two different (and often opposite) distributions $p(y|x)$ and $p(\bar{y}|x)$ for the same input $x$. For the vehicle following use-case presented here, these distributions represent the \textit{safe} action $a^s$ in current state $s$ and the \textit{unsafe} action $a^c$ in the same state. Simply, the safe action $a^s$ is learned from an expert demonstration of the task, whilst the unsafe action is learned from a dataset of collisions and represents an action likely to lead to a collision in the short term. Therefore, for any given state $s_t$, the AMDN models the probability of safe action $p(a_t^s|s_t)$ and the probability of action leading to collisions $p(a_t^c|s_t)$. For the vehicle following use-case presented here, unimodal policies are sufficient to solve the task, and since we are not using the different distributions to estimate uncertainty \cite{choi2018uncertainty} or multi-modal policies \cite{yan2019learning}, we represent each distribution by a unimodal Gaussian distribution, such that $p(a_t^s|s_t) = \mathcal{N}^s(\mu^s, {\sigma^s}^2)$ and $p(a_t^c|s_t) = \mathcal{N}^c(\mu^c, {\sigma^c}^2)$. However, it is worth noting that it would be straightforward to extend our approach to use mixtures of density functions for each distribution, should multi-modal representations be desired for the given task.

This additional information about safe and unsafe actions can then be employed to learn a safer control policy, by penalising the safe action distribution when it becomes similar to the unsafe action distribution. To do this, a similarity measure between the two distributions is required. A common statistical measure of the difference between two distributions is the Kullback-Leibler (KL) divergence $D_{KL}$ \cite{kullback1951information}. For two distributions the KL-divergence is given as:

\begin{equation}
D_{KL}(p(a^s_t|s_t)||p(a^c_t|s_t)) = \sum p(a^s_t|s_t) log\left( \frac{p(a^s_t|s_t)}{p(a^c_t|s_t)} \right)
\end{equation}

\begin{figure}
	\centering
	\includegraphics[width=0.5\textwidth]{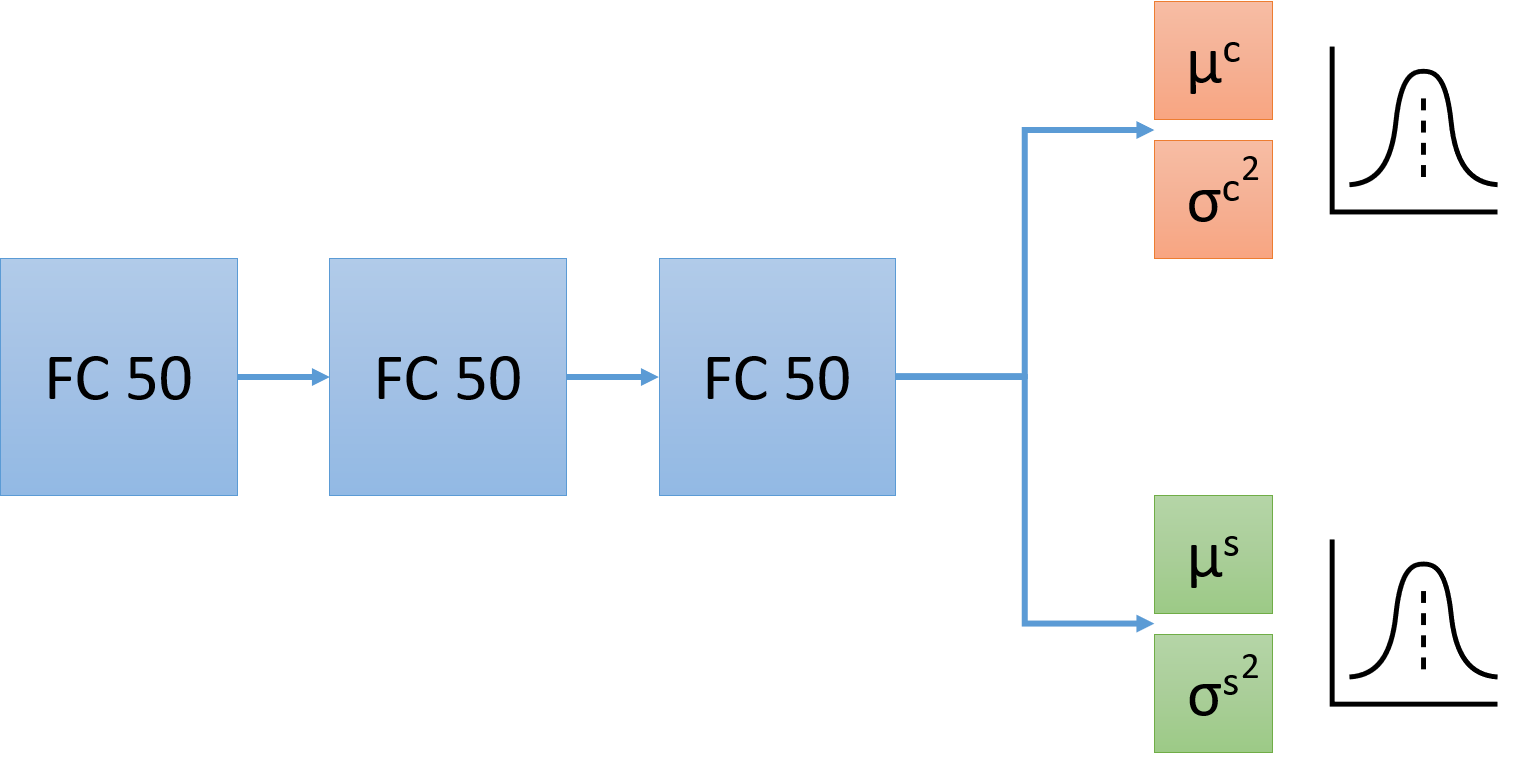}
	\caption{Proposed Adversarial Mixture Density Network architecture.}
	\label{fig_netarch}
\end{figure}

The AMDN is represented by a feedforward neural network, with three hidden layers of 50 neurons each, followed by the output layers which represent the mean and variance of the two Gaussian distributions $\mu^s$, ${\sigma^s}^2$, $\mu^c$, and ${\sigma^c}^2$, as shown in Fig. \ref{fig_netarch}. The two distributions are each trained on their respective datasets, $\mathcal{D}^e = (s_t^s, \hat{a}_t^s)$ and $\mathcal{D}^c = (s_t^c, \hat{a}_t^c)$. Both datasets are split into training and validation splits by a 80\%/20\% ratio, while further closed-loop testing within the simulator is used to test the robustness of the final trained policies. For both datasets, the action of the agent $a_t$ is represented by a single value in the range [-1, 1], which represents the use of the gas and brake pedals of the host vehicle, where positive values indicate the use of the gas pedal and negative values the brake pedal. The observed states $s_t$ consist of the host vehicle velocity $v$, relative velocity to the lead vehicle $v_{rel}$, and time headway to the lead vehicle $t_h$. The time headway is a measure of inter-vehicle distance in time, given by:
\begin{equation}
t_h = \frac{x_{rel}}{v}
\end{equation}
where $x_{rel}$ is the relative distance between two vehicles in m, and $v$ is the velocity of the host vehicle in m/s.

The expert dataset $\mathcal{D}^e$ is the dataset presented in \cite{kuutti2019safe}, which consists of 2 hours of driving data by the IPG CarMaker Simulator's \cite{IPG2017} default driver (IPG Driver) driving in different highway driving scenarios while aiming to maintain a time headway of 2s. The dataset consists of a total of 375,000 observation-action pairs. 

The collision dataset $\mathcal{D}^c$ is a dataset collected by utilising the adversarial testing framework in \cite{kuutti2020training} and stress testing a vehicle following controller. The testing framework uses adversarial agents to evaluate the target control policy. The adversarial agents are trained through reinforcement learning, with the aim of learning behaviours which cause the follower vehicle to crash into it. To ensure the collisions are preventable, and therefore expose a weakness in the target control policy, the actions and states of the adversaries are constrained based on minimum/maximum velocities and accelerations. The reward function of the adversary $r^A_t(s^A_t, a^A_t)$ is based on the inverse time headway to the follower vehicle, given by
\begin{equation}
r^A_t = min(\frac{1}{t_h}, 100)
\end{equation}
Where $r^A_t$ is the adversary's reward at time-step $t$, and the reward is capped at 100 to avoid the reward function tending towards infinity as the headway reaches 0. To collect the dataset, the adversary was deployed in different velocity ranges in highway driving, and the follower vehicle was controlled by an imitation learning agent trained on the  dataset $\mathcal{D}^e$. In this way, a dataset representing over 11,000 collision scenarios was collected. For each collision in the dataset, the 25 time-steps before the collision are taken for a total of 1s of driving data per collision and used for training, for a total of 275,000 observation-action in the dataset. 

We train the two distributions $\mathcal{N}^s$ and $\mathcal{N}^c$ with a Negative Log-Likelihood (NLL) loss, on the $\mathcal{D}^e$ and $\mathcal{D}^c$ datasets, respectively. The NLL loss $\mathcal{L}_{NLL}$ maximises the likelihood of given label actions $\hat{a}_t$ given an observation $s_t$:
\begin{equation}
\mathcal{L}_{NLL}(s_t, \hat{a}_t) = -log(\mathcal{N}(\hat{a}_t|s_t))
\end{equation}

In addition to each NLL loss, we also train the safe action distribution $\mathcal{N}^s$ on the KL-divergence loss. Since the aim is that the safe action distribution should avoid overlap with the actions that could lead to collisions, we maximise the KL-divergence between the two distributions. Therefore, the optimal parameters $\theta^*$ are obtained by the following three losses:
\begin{multline}
\theta^* = \arg \min_{\theta^{s}} \sum_{t} \mathcal{L}^{s}_{NLL}(\mathcal{N}^s, \hat{a}_t|s_t \sim \mathcal{D}^e) \\
+ \mathcal{L}^{c}_{NLL}(\mathcal{N}^c, \hat{a}_t|s_t \sim \mathcal{D}^c) 
- D_{KL}(\mathcal{N}^s| \mathcal{N}^c, s_t \sim \mathcal{D}^c)
\end{multline}

The hyperparameters used for training the AMDN are given in Table \ref{tbl_hyper}. Each neuron in the hidden layer uses the Rectified Linear Unit (ReLU) activation function, whilst each $\mu$ output uses a tanh activation function, and each $\sigma^2$ output uses a Non-negative Exponential Linear Unit (NNeLU) activation. Each loss $\mathcal{L}_{NLL}^s$, $\mathcal{L}_{NLL}^c$, $D_{KL}$ uses learning rates $\eta_{s}$, $\eta_{c}$, $\eta_{KL}$, respectively. All networks were trained using the Adam \cite{kingma2014adam} optimiser.

\begin{table}
	\renewcommand{\arraystretch}{1.25}
	\caption{AMDN hyperparameters.}
	\label{tbl_hyper}
	\centering
	\begin{tabular}{ m{5cm}  m{1cm} }
		\hline
		\hline
		\textbf{Parameter} & \textbf{Value} \\
		\hline
		Safe action learning rate, \textit{$\eta_{s}$} & 1x10\textsuperscript{-4} \\
		Unsafe action learning rate, \textit{$\eta_{c}$} & 1x10\textsuperscript{-5} \\
		KL-divergence learning rate, \textit{$\eta_{KL}$} & 1x10\textsuperscript{-9} \\
		Hidden layers & 3 \\
		Hidden neurons per layer & 50 \\
		Batch size & 100 \\
		Training steps & 1x10\textsuperscript{6} \\
		\hline
		\hline
	\end{tabular}
	
\end{table}

\section{Results}

\begin{table*}
	\renewcommand{\arraystretch}{1.2}
	\caption{Testing of learned control policies under Natural (Nat.) and Adversarial (Adv.) Testing frameworks, with baseline comparison including Imitation Learning with Feed-Forward Network, a standard Mixture Density Network, and different versions of Adversarial Mixture Density Networks.}
	\label{tbl_simresults}
	\centering
	\resizebox{1.0\linewidth}{!}{
		\begin{tabular}{c c c c c c c }
			\hline
			\hline
			\textbf{Testing Framework} & \textbf{Parameter} & 
			\begin{tabular}{@{}c@{}}\textbf{FFN} \\  \scriptsize\textbf{\cite{kuutti2019safe}} \end{tabular}  & 
			\begin{tabular}{@{}c@{}}\textbf{MDN} \\  \scriptsize\textbf{} \end{tabular} & 
			\begin{tabular}{@{}c@{}}\textbf{AMDN} \\  \scriptsize\textbf{(w/o $D_{KL}$)} \end{tabular} & 
			\begin{tabular}{@{}c@{}}\textbf{AMDN} \\  \scriptsize\textbf{(sampling)} \end{tabular} & 
			\begin{tabular}{@{}c@{}}\textbf{AMDN} \\  \scriptsize\textbf{} \end{tabular} \\
			\hline
			\multirow{7}{*}{Nat. Testing} 
			& min. x\textsubscript{rel} [m] & 23.84 & 0.00 & 0.00 & 7.66 & 11.34 \\
			& mean x\textsubscript{rel} [m] & 57.37 & 54.57 & 55.75 & 57.55 & 56.29 \\
			& max. v\textsubscript{rel} [m/s] & 8.88 & 14.56 & 14.21 & 11.79 & 10.93 \\
			& mean v\textsubscript{rel} [m/s] & 0.0197 & 0.0126 & 0.0138 & 0.0219 & 0.0224 \\
			& min. t\textsubscript{h} [s] & 1.74 & 0.00 & 0.00 & 0.90 & 1.13 \\
			& mean t\textsubscript{h} [s] & 1.99 & 1.90 & 1.94 & 2.00 & 1.95 \\
			& collisions & 0 & 1 & 2 & 0 & 0  \\
			\hline
			\multirow{2}{*}{Adv. Testing}
			& collisions against adversaries & 800 & 63 & 51 & 1 & 0  \\ 
			& episodes until collision & 245 & 684 & 1998 & 1974 & -  \\
			\hline
			\hline
		\end{tabular}
	}
	
\end{table*}

We test the effectiveness of the proposed AMDN approach in two different evaluation frameworks. \textit{Naturalistic testing} tests the control policy in common driving scenarios which represent the typical scenarios encountered when driving on highways. These tests are completed in 5-minute episodes with pre-defined lead vehicle trajectories in the IPG CarMaker Simulator. The types of lead vehicle trajectories seen are similar to those seen in the expert driver dataset $\mathcal{D}^e$. For each model, we test the control policy in 120 different scenarios, which totals up to 10 hours of driving. In contrast, \textit{adversarial testing} utilises the adversarial agents presented in \cite{kuutti2020training} to create safety-critical scenarios. The aim of these agents is to act in such a way that the vehicle follower should collide into them. However, the actions of the adversarial agents are limited such that any collisions would have been preventable, and therefore represent a fault in the vehicle follower's control policy. We utilise the lead vehicle velocity and acceleration limits which were shown to be most effective for creating collisions in highway driving in \cite{kuutti2020training}, such that the velocity and acceleration are limited to $v_{lead} \in [12, 30]$ m/s and $a_{lead} \in [-6, 2]$ m/s\textsuperscript{2}, respectively. For each model tested, we train 5 adversarial agents for up to 2,500 episodes, and then report the number of collisions averaged over the 5 training runs. It is worth noting that, while the adversarial testing framework is the same as the one used to collect the dataset $\mathcal{D}^c$, each of the adversaries used in the testing in this Section is randomly initialised and trained specifically against one network only. If the same adversarial agents that were used to collect the dataset would be used for validation purposes, the agents could encounter similar adversarial behaviour seen during training. Instead, these adversarial agents are not the same agents as those that collected the dataset $\mathcal{D}^c$, and the adversarial testing demonstrates how easy it is for new adversaries to find control strategies that can exploit weaknesses in the target policies. For both types of testing, at the beginning of each test episode, a coefficient of friction between 0.4 and 1.0 is randomly chosen.

We compare the AMDN to two different baselines, FFN is the Feed-Forward Network based Imitation Learning control policy presented in \cite{kuutti2019safe} and MDN is a standard Mixture Density Network with $M = 1$, which outputs a single Gaussian Distribution trained on dataset $\mathcal{D}^e$. During inference, all MDN and AMDN variants, unless otherwise stated, use the output $\mu^s$ as the pedal action, since during testing it was shown that this significantly improves the stability of the vehicle control policy. To compare this inference strategy to sampling from the distribution $\mathcal{N}^s$, we provide comparison to the AMDN (sampling) model which samples action from the Gaussian distribution during inference. We also compare the model to the AMDN without the KL-divergence loss, which is denoted by AMDN (w/o $D_{KL}$).

\begin{figure}[t]
	\centering
	\includegraphics[width=0.5\textwidth]{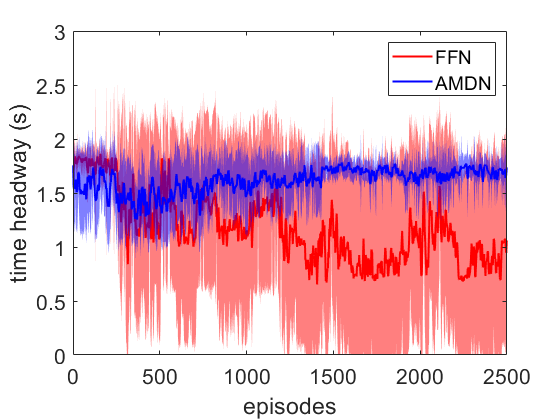}
	\caption{Minimum episode headway during Adversarial Testing. Averaged over 5 training runs, with standard deviation shown in the shaded region.}
	\label{fig_advtest}
\end{figure}

The summary of the results from the naturalistic testing are provided in the top half of Table \ref{tbl_simresults}, while the results of the adversarial testing are shown in lower half of the table. Comparing the AMDN to the AMDN (sampling) model shows that using the $\mu^s$ as the action during deployment is a better inference strategy, as the AMDN policy using $\mu^s$ as the action during inference maintains a safer distance from the lead vehicle. Moreover, when sampling the pedal action from the $\mathcal{N}^s$ distribution, the variance of the sampled actions creates jerky driving trajectories with rapid accelerations and decelerations, which can present safety and passenger comfort problems when driving. However, it is still beneficial to model the two Gaussian distributions during training, as this allows the use of the KL-divergence loss on the unlabelled collision dataset. The benefit of the $D_{KL}$ loss can be seen by comparing the performance of the AMDN to the AMDN (w/o $D_{KL}$) model. Not only does the AMDN (w/o $D_{KL}$) collide 2 times in the naturalistic driving, but it shows significantly greater susceptibility to adversarial testing as shown by its 51 collisions, demonstrating the significant benefit of using the KL-divergence loss during training. Moreover, comparing the AMDN to the standard MDN also shows the benefit of the proposed approach, as the AMDN shows significantly better driving performance in both the naturalistic and adversarial testing compared to MDN. Of the models compared here, the only two models that can drive without collisions in naturalistic driving are the FFN and AMDN. The results demonstrate that while the two models can drive safely without collisions, the FFN performs slightly better in naturalistic driving as it has slightly higher minimum headway at 1.74s compared to 1.13s and a minimum distance of 23.84m compared to 11.34m. However, on average the models show no significant difference as the FFN has a mean headway of 1.99s compared to 1.95s for AMDN, and the mean distances are 57.37m compared to 56.29m. However, comparing the FFN and AMDN during adversarial testing shows one of the main benefits of the AMDN approach. The FFN shows significant vulnerability when facing unpredictable driving behaviour different from its training distribution, as demonstrated by its average number of 800 collisions in adversarial testing, whilst the AMDN can still drive without collisions even in the presence of adversarial agents that attempt to deliberately cause collisions. The minimum headway during adversarial testing can be seen in Fig. \ref{fig_advtest}, where the results are averaged over the 5 adversarial agents trained against each policy. As can be seen the minimum headway for the FFN policy reduces over the training time with increasing variance as the adversarial agents learn to exploit the FFN model. In comparison, the minimum headway for AMDN remains over 1s without reducing significantly, suggesting the adversarial agents are struggling to learn how to exploit the AMDN thus demonstrating increased robustness to unpredictable and adversarial road users.

\begin{figure}
	\centering
	\includegraphics[width=0.45\textwidth]{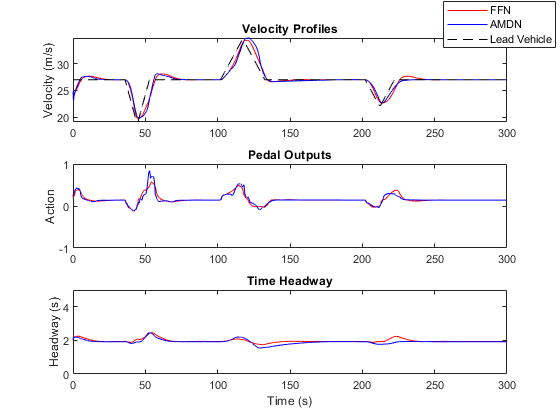}
	\caption{Example vehicle following scenario from naturalistic driving tests. Vehicle velocities (top), host vehicle pedal actions (middle), and relative time headway (bottom).}
	\label{fig_nattest1}
\end{figure}

\begin{figure}
	\centering
	\includegraphics[width=0.45\textwidth]{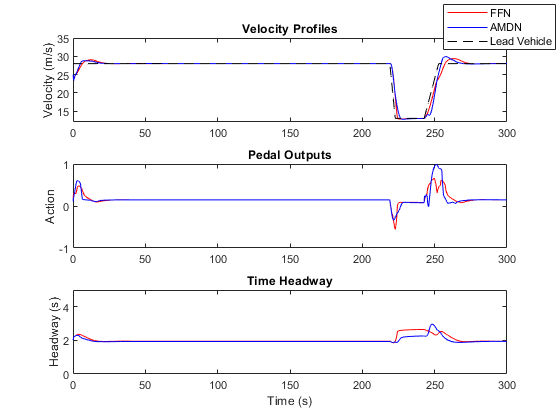}
	\caption{Example vehicle following scenario from naturalistic driving tests. Vehicle velocities (top), host vehicle pedal actions (middle), and relative time headway (bottom).}
	\label{fignattest2}
\end{figure}

\begin{figure}
	\centering
	\includegraphics[width=0.45\textwidth]{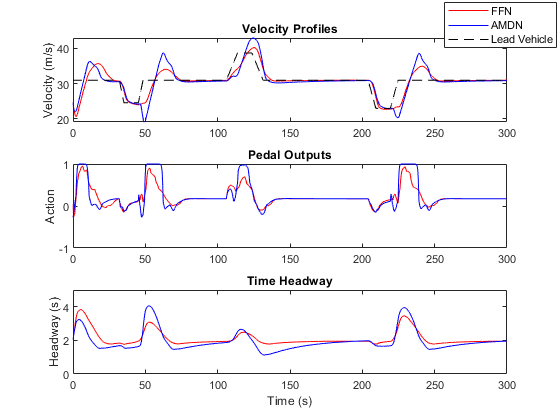}
	\caption{Example vehicle following scenario from naturalistic driving tests. Vehicle velocities (top), host vehicle pedal actions (middle), and relative time headway (bottom).}
	\label{fig_nattest3}
\end{figure}

In addition to the summaries of the different tests, we also show some example episodes in Fig. \ref{fig_nattest1} - \ref{fig_nattest3} comparing the learned driving policies by the FFN and AMDN models under naturalistic driving. The example scenario in Fig. \ref{fig_nattest1} shows the lead vehicle starting at a constant velocity, and then decelerating rapidly following by immediately accelerating and vice versa. We see that both the FFN and AMDN models can follow this type of trajectories easily without straying far from the target headway of 2s. The control policies' response to an emergency braking manoeuvre by the lead vehicle is shown in Fig. \ref{fignattest2}. The scenario occurs in low friction conditions with a road friction coefficient of 0.475. At $t$ =  224s, the lead vehicle suddenly decelerates at 4m/s\textsuperscript{2}, forcing the host vehicle to brake in response. The vehicle then drives at 13m/s for a period, followed by accelerating at 1.5m/s\textsuperscript{2} back to its original velocity. As can be seen in Fig. \ref{fignattest2}, both the FFN and AMDN models can adequately respond to this harsh braking in low friction conditions, and avoid any potential collision without bringing the vehicle too close to the lead vehicle. To demonstrate the worst case behaviour by the AMDN, in Fig. \ref{fig_nattest3} the test scenario in which the AMDN control policy demonstrated its lowest headway in naturalistic testing is presented. This scenario includes multiple accelerations and decelerations by the lead vehicle. We can see that both vehicle followers again maintain a safe distance from the lead vehicle. However, at two points in the scenario we can see the lead vehicle decelerate to a lower velocity, and after some time accelerating back to its original velocity. As the lead vehicle starts to accelerate in this manoeuvre, we can see the AMDN decide to engage the brakes of the vehicle, followed by rapid acceleration to catch up to the lead vehicle. While this behaviour does not result in any safety-critical situation, this type of jerky manoeuvre could present comfort concerns for any passengers. However, these models were not trained explicitly to consider passenger comfort, therefore expanding this work to include considerations for passenger comfort could be included in future work.

\begin{figure}[t]
	\centering
	\includegraphics[width=0.45\textwidth]{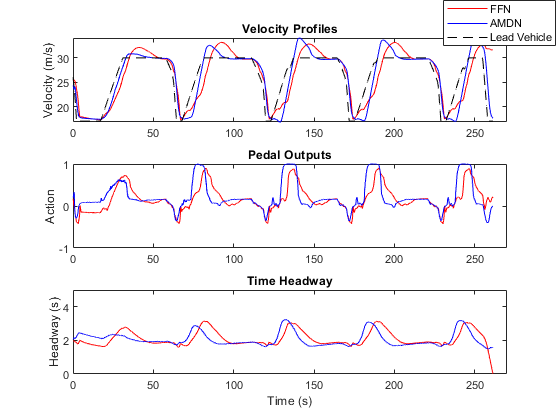}
	\caption{Example vehicle following scenario from adversarial testing. Vehicle velocities (top), host vehicle pedal actions (middle), and relative time headway (bottom).}
	\label{figadvtest1}
\end{figure}

\begin{figure}[t]
	\centering
	\includegraphics[width=0.45\textwidth]{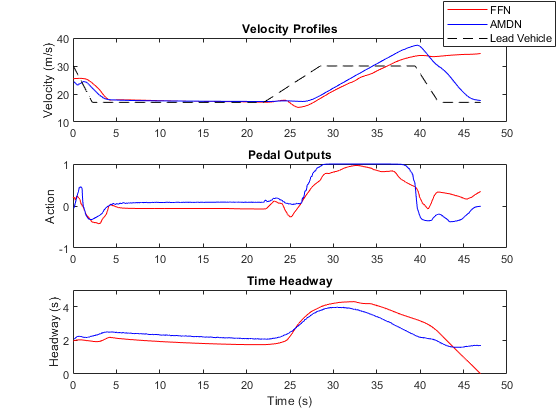}
	\caption{Example vehicle following scenario from adversarial testing. Vehicle velocities (top), host vehicle pedal actions (middle), and relative time headway (bottom).}
	\label{figadvtest2}
\end{figure}

Investigating the behaviour of the two models in the adversarial testing can provide further insight into the learned policies. An example collision episode from the adversarial testing of the FFN is shown in Fig. \ref{figadvtest1}. As can be seen, the adversarial lead vehicle keeps decelerating to a low velocity followed by accelerating to a high velocity, and then continues to repeat this manoeuvre. While the AMDN keeps a safe distance with a minimum headway of 1.48s, the FFN eventually fails to maintain a safe distance as it accelerates towards the lead vehicle to catch up but the lead vehicle suddenly breaks harshly and a collision occurs. Another example from the adversarial testing is seen in Fig. \ref{figadvtest2}, where again the adversarial lead vehicle starts from a low velocity and then rapidly accelerates to a high velocity. As the follower vehicle is accelerating to catch the lead vehicle, the adversary then breaks suddenly. However, this scenario shows that even when the lead vehicle is already decelerating rapidly, the FFN agent continues to accelerate because the vehicles have a large relative headway and the FFN is trying to reach a headway of 2s. Meanwhile, the AMDN agent responds to the rapid deceleration by the lead vehicle by also braking, even though the headway between the agents is above 2s. These results suggest that the FFN model, trained only on the expert demonstrated data where a 2s headway was always followed, considers the headway more important while the AMDN focuses on maintaining the same speed with the lead vehicle. This demonstrates a difference in the learned driving strategies of the two models, and also provides further insight into why the AMDN has lower minimum headways compared to FFN  in naturalistic driving, despite demonstrating better capabilities at avoiding collisions. Indeed, this could be a driving strategy it learned from the collision dataset $\mathcal{D}^c$, which helps the model be robust to these types of adversarial agents and maintain a collision free driving performance. 

\section{Conclusions}
In this paper, a technique for learning safe control policies for autonomous driving was presented. The proposed technique is based on Mixture Density Networks, where the model learns two separate action distributions. The first distribution is a safe action distribution learned from a dataset of expert demonstration, whilst the second is a unsafe action distribution learned from a dataset of collisions. By learning each of these distributions, the model then uses a KL loss, which penalises the safe action distribution for similarity to unsafe action distribution, using example states from the collision dataset. In this way, the model learns a more safe and robust control policy, which remains safe even in dangerous edge-cases. The proposed approach was demonstrated in a highway vehicle following use-cases, and the Adversarial Mixture Density Networks were compared to standard imitation learning with feed-forward networks and mixture density networks. The results demonstrate that the proposed approach results in a more safe control policy, which is better able to react to unpredictable and adversarial road users.



\section*{Acknowledgment}
This work was funded by the EPSRC under grant agreements (EP/R512217/1) and (EP/S016317/1).


\bibliographystyle{IEEEtran}
\bibliography{ref_amdn}

\end{document}